\documentclass[pdflatex,sn-apa]{sn-jnl}

\usepackage{graphicx}%
\usepackage{multirow}%
\usepackage{amsmath,amssymb,amsfonts}%
\usepackage{amsthm}%
\usepackage{mathrsfs}%
\usepackage[title]{appendix}%
\usepackage{xcolor}%
\usepackage{textcomp}%
\usepackage{manyfoot}%
\usepackage{booktabs}%
\usepackage{listings}%

\usepackage[cjk]{kotex}
\usepackage{arydshln}
\usepackage{setspace}
\usepackage{algorithmic,algorithm}
\usepackage{arydshln}

\usepackage{float}
\usepackage[caption = false]{subfig}

\usepackage{synttree}

\theoremstyle{thmstyleone}%
\theoremstyle{thmstyletwo}%
\theoremstyle{thmstylethree}%
\begin{document}

\title[Punctuation-aware binarization]{Punctuation-aware treebank tree binarization}

\author[1]{\fnm{Eitan} \sur{Klinger}}\email{ekling01@student.ubc.ca}
\equalcont{}
\author[1]{\fnm{Vivaan} \sur{Wadhwa}}\email{vivaanw@student.ubc.ca}
\equalcont{E. Klinger and V. Wadhwa contributed equally to this work.}
\author*[2]{\fnm{Jungyeul} \sur{Park}}\email{jungyeul.park@gmail.com}
\affil[1]{\orgdiv{Department of Computer Science}, \orgname{The University of British Columbia}, \orgaddress{
\country{Canada}}}
\affil[2]{\orgdiv{Culture Technology Research Institute}, \orgname{Korea Advanced Institute of Science \& Technology}, \orgaddress{
\country{South Korea}}}

\abstract{
This article presents a curated resource and evaluation suite for punctuation-aware treebank binarization. 
Standard binarization pipelines drop punctuation before head selection, which alters constituent shape and harms head-child identification. 
We release (1) a reproducible pipeline that preserves punctuation as sibling nodes prior to binarization, (2) derived artifacts and metadata (intermediate @X markers, reversibility signatures, alignment indices), and (3) an accompanying evaluation suite covering head-child prediction, round-trip reversibility, and structural compatibility with derivational resources (CCGbank). 
On the Penn Treebank, punctuation-aware preprocessing improves head prediction accuracy from 73.66\% (Collins rules) and 86.66\% (MLP) to 91.85\% with the same classifier, and achieves competitive alignment against CCGbank derivations. 
All code, configuration files, and documentation are released to enable replication and extension to other corpora. 
}


\maketitle

\section{Introduction}

Treebank-based parsers generally rely on binary branching. This design choice, though often treated as a technical prerequisite, has profound implications for how syntactic structures are represented and processed. Binarization enables chart-based algorithms such as CKY parsing \citep{cocke:1969,kasami:1966,younger:1967} and serves as the foundation for transition-based systems \citep{sagae-lavie-2005-classifier,zhu-etal-2013-fast}. In both cases, binary branching simplifies inference and parameter estimation by ensuring uniform rule arity. Yet converting multi-child nodes into binary form is not a neutral operation: it reshapes the structure of the source data, thereby affecting every downstream parser and grammar derived from it.

The conversion is typically guided by head percolation tables \citep{collins:1999}, which define how internal nodes and head–child relations are assigned during binarization. These head rules encode assumptions about syntactic headedness and child ordering, becoming a standard component in virtually all Penn Treebank (PTB) preprocessing pipelines. Over the years, this procedure has propagated into diverse frameworks-from generative phrase-structure grammars to discriminative transition systems-forming the structural backbone of many syntactic resources.

A key feature of this legacy is the systematic \emph{exclusion of punctuation} from internal tree structures. Far from being an oversight, this practice originates from early evaluation design. In the PARSEVAL framework \citep{black-etal-1991-procedure}, punctuation was intentionally ignored to achieve reproducible bracket scoring across tokenization variants. The implementation of \texttt{evalb}, which became the de facto standard for evaluating constituency parsers, formalized this decision. Later, the Collins parser \citep{collins:1999} extended the same convention to head-rule definitions, treating punctuation as non-syntactic and removing it before binarization. These design choices were driven by pragmatic goals-computational simplicity, comparability, and evaluation stability-not by any claim that punctuation lacks grammatical relevance. However, the cumulative effect was to institutionalize punctuation exclusion across all PTB-derived resources and parser training pipelines.

This convention, while convenient, has long-term structural consequences. Punctuation encodes boundary cues that signal clause completion, coordination, and apposition. When such tokens are removed before binarization, constituent structures are forced into artificial configurations that misrepresent original annotations. The resulting trees remain formally valid but are no longer structurally faithful: they distort head–child relations and weaken correspondence with derivational formalisms such as CCGbank \citep{hockenmaier-steedman-2007-ccgbank}, where punctuation explicitly governs combinatory scope. These discrepancies, though subtle at the sentence level, accumulate across thousands of trees, introducing inconsistencies that propagate into downstream analyses and model training.

\begin{figure}
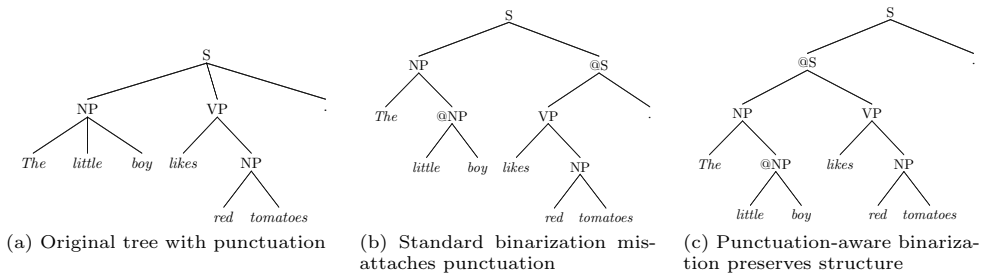

\centering
\subfloat[Original tree with punctuation \label{original-tree}]{
\resizebox{.32\textwidth}{!}{
\synttree
[S [NP [\textit{The}] [\textit{little}] [\textit{boy}]]
[VP [\textit{likes}] [NP [\textit{red}] [\textit{tomatoes}]]]
[$\cdot$]]
}}
\quad
\subfloat[Standard binarization misattaches punctuation \label{wrong-binary}]{
\resizebox{.29\textwidth}{!}{
\synttree
[S [NP [\textit{The}] [@NP [\textit{little}] [\textit{boy}]]]
[@S [VP [\textit{likes}] [NP [\textit{red}] [\textit{tomatoes}]]] [$\cdot$]]]
}}
\quad
\subfloat[Punctuation-aware binarization preserves structure \label{correct-binary}]{
\resizebox{.29\textwidth}{!}{
\synttree
[S [@S [NP [\textit{The}] [@NP [\textit{little}] [\textit{boy}]]]
[VP [\textit{likes}] [NP [\textit{red}] [\textit{tomatoes}]]]] [$\cdot$]]
}}
\caption{Excluding punctuation during binarization distorts constituent structure (b), corrected by including punctuation as a syntactic sibling (c).}
\label{treebank-binary-tree-example}
\end{figure}

Figure~\ref{treebank-binary-tree-example}, adapted from \citet{liu-zhang-2017-order}, illustrates a typical case. The original tree (Figure~\ref{original-tree}) integrates the final period as part of the surface structure. In standard binarization (Figure~\ref{wrong-binary}), punctuation is dropped and later reattached to the root, yielding a right-branching VP structure absent in the original annotation. Our punctuation-aware binarization (Figure~\ref{correct-binary}) instead retains the punctuation as a sibling node before binarization, preserving both syntactic closure and clause boundaries. The resulting binary tree remains compatible with existing parsers but more accurately reflects the intended structure.

Retaining punctuation is therefore less about parser performance than about \emph{resource reliability}. Punctuation-aware binarization restores the fidelity between surface tokens and syntactic annotation, ensuring that binary trees remain linguistically interpretable and reproducible across formal frameworks. This reproducibility is essential for maintaining consistency among interdependent resources such as CCGbank and TAGbank, whose derivations rely on accurate boundary information from PTB.

This paper introduces a reproducible, reversible binarization resource that preserves punctuation as structural information. The proposed procedure yields binary trees that are fully compatible with existing treebanks, require no new annotation, and support faithful round-trip conversion. By aligning formal representation with linguistic structure, this work contributes to the methodological transparency of treebank preprocessing and facilitates standardization across future resources. 

We evaluate this approach along three dimensions: 
(1) {Hypothesis 1 (Structural fidelity):} retaining punctuation reduces distortion in constituent configuration; 
(2) {Hypothesis 2 (Structural predictability):} punctuation provides disambiguating cues that improve head–child identification accuracy; and 
(3) {Hypothesis 3 (Reproducibility):} explicit marking guarantees fully reversible and lossless binarization.  

These results collectively show that punctuation should not be regarded as a peripheral artifact but as an integral component of syntactic structure. Its inclusion enhances the reliability, transparency, and interpretability of treebank resources and provides a consistent basis for future extensions across languages and formalisms.

\section{Methodology: punctuation-aware binarization}

Punctuation marks such as commas, periods, and quotation marks occur in almost every sentence in the Penn Treebank, yet they are normally excluded from the internal structure of parse trees. Conventional binarization procedures treat punctuation as non-syntactic, remove it before structural conversion, and then reattach it as isolated terminal leaves. Although this practice originated from efficiency considerations and the evaluation conventions of PARSEVAL \citep{black-etal-1991-procedure}, it distorts the syntactic organization of clauses and appositives and disrupts the relation between surface form and structural representation.

This section defines a deterministic procedure that preserves punctuation within the tree structure while maintaining binary branching. Each punctuation mark is treated as a sibling node attached to its syntactic neighbor before binarization. The procedure operates in linear time with respect to the number of tokens.

\subsection{Transformation procedure}

The algorithm, denoted \( f_{\text{punct}}(T) \), converts an original constituency tree \(T\) into its punctuation-aware binary form. The inverse transformation \( f^{-1}_{\text{punct}} \) restores the original tree, satisfying the identity property  
\( f^{-1}_{\text{punct}}(f_{\text{punct}}(T)) = T \).

\begin{algorithm}[!ht]
\caption{Punctuation-aware binarization}
\begin{algorithmic}[1]
\STATE \textbf{Input:} Constituency tree $T$
\STATE \textbf{Output:} Binarized tree $T' = f_{\text{punct}}(T)$
\FOR{each node $X$ in $T$ in depth-first order}
  \FOR{each child $c_i$ of $X$}
    \IF{$c_i$ is punctuation}
      \STATE Determine attachment direction
      \IF{$c_i$ precedes a non-punctuation child}
        \STATE $\textit{dir} \gets \text{RIGHT}$
      \ELSE
        \STATE $\textit{dir} \gets \text{LEFT}$
      \ENDIF
      \STATE Insert intermediate node \texttt{@X} with flag \textit{dir}
      \STATE Attach $c_i$ as a sibling of its neighboring constituent
      \STATE Record positional flag and structural signature for reversibility
    \ENDIF
  \ENDFOR
\ENDFOR
\STATE \textbf{return} $T'$
\end{algorithmic}
\end{algorithm}

Each punctuation token is processed independently, so sequences such as commas followed by quotation marks are attached in left-to-right order. If no punctuation occurs within a constituent, the original structure is left unchanged. The reversibility signature uniquely identifies the inserted intermediate nodes (\texttt{@X}) so that they can be removed deterministically during the inverse transformation.

\subsection{Structural representation}

The resulting local configuration has the general form
\[
(\textsc{X} \rightarrow (\textsc{@X}\ nt_0 \ldots nt_n)\ \blacktriangleleft\texttt{punct}),
\]
where \texttt{nt} represents a non-terminal child, \texttt{@X} an intermediate node created during binarization, and $\blacktriangleleft$ marks right-attaching punctuation such as commas or periods. The dual symbol $\blacktriangleright$ marks left-attaching punctuation such as opening quotation marks or parentheses. These operators preserve the original adjacency relation of punctuation to the relevant phrase boundary.

Figure~\ref{punct-rules} summarizes the restructuring heuristics. Each rule represents a local adjustment that preserves constituent integrity while maintaining punctuation's syntactic location.

\begin{figure}[!ht]
\begin{center}
\resizebox{\textwidth}{!}{
\begin{tabular}{|r l c l|}\hline
{r}$_1$ & 
($_{X}$ $\cdots$ $nt_{i}$ $\texttt{punct}\blacktriangleright$ $nt_{j}$ $\cdots$ ) 
&  $\rightarrow$ & 
($_{X}$ $\cdots$ $nt_{i}$ ($_{@X}$ $\texttt{punct}\blacktriangleright$  ($_{@X}$ $nt_{j}$ $\cdots$ ) ) ) \\
\hdashline

{r}$_2$ & 
($_{X}$ $\cdots$ $nt_{i}$ $\blacktriangleleft\texttt{punct}$ $nt_{j}$ $\cdots$ ) 
&  $\rightarrow$  & 
($_{X}$ ($_{@X}$ ($_{@X}$ $\cdots$ $nt_{i}$ ) $\blacktriangleleft\texttt{punct}$ ) $nt_{j}$  $\cdots$  ) 
\\
\hdashline

{r}$_3$& 
($_{X}$ $\cdots$ $nt_{i}$ $\texttt{punct}\blacktriangleright$ $nt_{j}$ $\cdots$ $\blacktriangleleft\texttt{punct}$ $nt_{k}$ $\cdots$ )
&  $\rightarrow$ & 
($_{X}$ $\cdots$ $nt_{i}$ ($_{@X}$ ($_{@X}$ $\texttt{punct}\blacktriangleright$ ($_{@X}$ $nt_{j}$ $\cdots$ ) ) $\blacktriangleleft\texttt{punct}$ ) $nt_{k}$ $\cdots$ )
\\
\hline
\end{tabular}
}
\end{center}
\caption{Restructuring rules applied before binarization to attach punctuation as syntactic siblings.
$\blacktriangleright$ indicates left-attaching punctuation (e.g., opening quotation marks or parentheses), and $\blacktriangleleft$ indicates right-attaching punctuation (e.g., commas or periods).}
\label{punct-rules}
\end{figure}

\subsection{Example}

\begin{figure}[!ht]
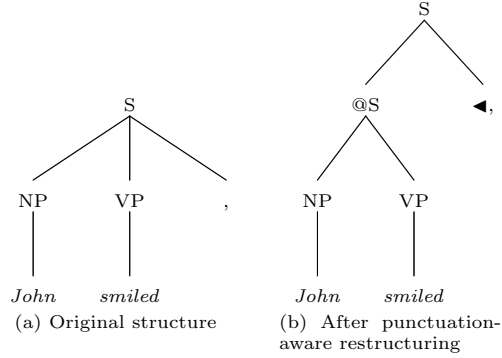

\centering
\footnotesize
\subfloat[Original structure]{
{
\synttree
[S [NP [\textit{John}]] [VP [\textit{smiled}]] [,]]
}}
\qquad
\subfloat[After punctuation-aware restructuring]{
{
\synttree
[S [@S [NP [\textit{John}]] [VP [\textit{smiled}]]] [$\blacktriangleleft$,]]
}}
\caption{Example of punctuation attachment and @X node marking.}
\label{punct-example}
\end{figure}

Figure~\ref{punct-example} illustrates a simplified before-and-after transformation.  
In the original structure, punctuation is external to the clause; after restructuring, the comma is retained as a sibling of the phrase it delimits, and the intermediate node \texttt{@S} carries the positional flag \(\blacktriangleleft\).

\subsection{Properties and reversibility}

Because every intermediate node is tagged with a unique positional flag and structural signature, the transformation is fully reversible. The inverse function \( f^{-1}_{\text{punct}} \) removes all \texttt{@X} nodes and restores the original tree in linear time. Formally, the pair \((f_{\text{punct}}, f^{-1}_{\text{punct}})\) defines a bijection between original and binarized trees under the same tokenization.

\subsection{Generality and adaptation}

Although the procedure was first implemented for the English Penn Treebank, its design is language-independent.  
Every written language employs a finite inventory of punctuation marks, each with predictable attachment behavior that can be specified in a configuration file.  
By redefining these attachment rules and punctuation types, the same binarization framework can be applied to any constituency-based treebank.

Languages differ in both the form and placement of punctuation.  
For example, Spanish and Galician use paired inverted marks ({¿}, {¡}) at clause openings, while East Asian scripts such as Chinese and Japanese employ full-width variants of commas and quotation marks that appear as distinct tokens.  
Other languages introduce mid-clause separators, paired dashes, or multiple quotation systems.  
Despite this variation, each punctuation mark can be formally categorized as either left-attaching or right-attaching, or as a symmetric pair whose orientation is determined by context.

The implementation therefore provides a user-defined punctuation map that enumerates all symbols and their corresponding attachment rules.  
This configuration enables consistent treatment of punctuation across different treebanks without changing the algorithm itself.  
Because the number of punctuation types in any language is finite, complete coverage can be achieved through a closed and easily extensible specification.



\section{Empirical evaluation of punctuation-aware binarization}
This section evaluates the empirical impact of punctuation-aware binarization on structural learning and representational fidelity.
We examine how preserving punctuation influences head-child identification accuracy, ensures reversibility of tree reconstruction, and improves alignment with derivation-based grammatical frameworks.
These evaluations quantify the syntactic and practical benefits of incorporating punctuation into binarized treebank structures.

\subsection{Head–child prediction}

To evaluate the effect of punctuation-aware binarization on structural reliability, we align the dependency-converted version of the Penn Treebank (PTB) \citep{de-marneffe-etal-2006-generating} with its original constituency structures \citep{marcus-etal-1993-building}.  
Each dependency head is treated as the corresponding constituent head in the aligned tree.  
Because of tokenization mismatches—primarily involving punctuation and hyphenated forms such as \textit{third - quarter} versus \textit{third-quarter}—84.3\% of sentences align perfectly and are used for evaluation.  
Unaligned sentences are excluded so that results reflect structural rather than tokenization differences.

We adopt the standard PTB split: Sections 02–21 for training, Section 22 for development, and Section 23 for testing.  
Three head-finding approaches are compared:  
(1) the rule-based \textsc{Collins} head percolation table \citep{collins:1999};  
(2) a multilayer perceptron classifier (\textsc{MLP}) trained on the original punctuation-stripped trees; and  
(3) the same classifier trained on trees modified by the punctuation-aware restructuring (\textsc{MLP+Punct}).  

The \textsc{MLP} model consists of two hidden layers with ReLU activation and is optimized with Adam for ten epochs with early stopping on the development set.  
Input features include the parent constituent label, child categories, linear order, and functional tags.  
In the punctuation-aware variant, an additional binary feature indicates whether a child node is adjacent to punctuation and whether it follows or precedes it.  
All neural experiments are run with five random seeds, and results are reported as mean accuracy with standard deviation to ensure statistical robustness.

Head–child prediction serves as a proxy for measuring the structural clarity of binarized trees.  
If punctuation-aware restructuring improves the accuracy of identifying a constituent’s head, it implies that punctuation contributes disambiguating cues that clarify local attachment decisions.  
This evaluation therefore reflects the linguistic fidelity of binarization rather than parser performance.

Table~\ref{find-head-results} reports head–child prediction accuracy on the Penn Treebank test set (Section~23).  
The learning-based classifier substantially outperforms the rule-based baseline, improving accuracy from 73.66\% with the Collins head rules to 86.66\% with the MLP model trained on punctuation-stripped trees.  
When punctuation is retained during preprocessing, accuracy increases further to 91.85\%.  
This finding supports the hypothesis that punctuation processing improves the accuracy of head–child identification.  
When punctuation is excluded, the MLP trained on the original trees achieves 86.66\%.  
After processing punctuation first and then applying head selection to the remaining constituents, accuracy increases to 91.85\%.  
This shows that handling punctuation separately before head prediction yields more reliable binarized structures.

\begin{table}[!ht]
\centering
\caption{Head–child prediction accuracy on the Penn Treebank (Section~23) using rule-based and learned models, with and without punctuation-aware preprocessing.}
\label{find-head-results}
\begin{tabular}{lccc}
\toprule
Method & Collins & MLP & MLP + Punct \\
\midrule
Accuracy (\%) & 73.66 & 86.66 & \textbf{91.85} \\
\bottomrule
\end{tabular}
\end{table}

\subsection{Reversibility and structural integrity}

The restructuring algorithm is fully reversible by design.  
During binarization, each intermediate node introduced to enforce binary branching is labeled \texttt{@X} and assigned a positional flag that records its local attachment direction.  
These flags act as unique structural signatures, allowing each auxiliary node to be deterministically removed during the inverse operation.  
Formally, for any tree $T$, the transformation $f_{\text{punct}}(T)$ produces a binary tree such that the inverse mapping satisfies
\[
f^{-1}_{\text{punct}}(f_{\text{punct}}(T)) = T
\]
This identity holds because the mapping is bijective: every inserted \texttt{@X} node and flag pair corresponds to a single structural modification that can be exactly reversed.

Reversibility was confirmed empirically on the evaluated portion of the Penn Treebank.  
All transformed trees were restored to their original configurations without loss of constituent labels, spans, or punctuation placement.  
This property ensures that the binarization process is lossless and reproducible, allowing consistent reuse of the same linguistic data for parser training, evaluation, or further resource conversion without altering the underlying annotation.

\subsection{Alignment with derivation-based schemes}

We assess the structural correspondence between our punctuation-aware binary trees and the derivations in CCGbank \citep{hockenmaier-steedman-2007-ccgbank}, using \texttt{jp-evalb} \citep{jo-etal-2024-novel}, which explicitly retains punctuation during evaluation. On Section~23 of the Penn Treebank, our method achieves an F1 score of 76.07\%, with recall of 75.17\% and precision of 76.99\%. CCGbank provides derivations for 2,407 of the 2,416 original sentences, but after enforcing exact alignment of terminal nodes, 1,858 sentences remain due to tokenization discrepancies involving punctuation and quotation marks. To ensure structural comparability, we simplified both representations by replacing all nonterminal labels with \texttt{nt}, preserving only POS–word pairs at the terminals, and removing type-raising unary rules in CCGbank.

Many of the remaining mismatches arise from CCGbank's handling of commas, which are promoted to top-level constituents in appositive contexts \citep[p.~23]{hockenmaier-steedman-2005-ccgbank-manual}. For instance, in \textit{the CEO, John Smith $\cdots$}, our trees attach the comma to the preceding noun phrase (\textit{the CEO $\blacktriangleleft$,} $\cdots$) to maintain local constituency, whereas CCGbank attaches it to the appositive phrase (\textit{$\cdots$ ,$\blacktriangleright$ John Smith} $\cdots$), assigning it a role analogous to coordination and structurally promoting it above the appositive.

These differences reflect not annotation inconsistencies but contrasting grammatical assumptions. CCG, by design, enforces binary branching and treats punctuation as structurally independent, often assigning it the same category used in coordination without full combinatory semantics. Our binarized trees, in contrast, incorporate punctuation directly into constituent structure, preserving surface adjacency cues and improving alignment in clause-boundary and sentence-final constructions. Some divergence persists where CCGbank simplifies appositives by promoting commas to top-level nodes.

Derivational frameworks such as Tree-Adjoining Grammar (TAG) similarly treat punctuation as a syntactically meaningful marker, typically realized through adjunction to indicate clause or phrase boundaries. Although our restructuring is not tied to a specific formalism, it brings treebank-based constituency representations closer to such derivation-oriented treatments by embedding the boundary-marking function of punctuation within the tree structure.

This alignment analysis demonstrates that punctuation-aware binarization not only enhances internal structural coherence but also increases interoperability with grammar-driven syntactic resources, thereby supporting cross-formalism comparisons and resource integration.

\subsection{Summary of findings}

Our empirical analyses demonstrate that punctuation-aware binarization enhances syntactic learning, ensures full reversibility of treebank transformations, and yields structures that better align with derivational formalisms such as CCG. Preserving punctuation during preprocessing recovers structural cues that are typically lost in conventional binarization pipelines, leading to measurable gains in both head-child prediction accuracy and structural comparability with grammar-based resources. These findings underscore the importance of treating punctuation as an integral component of syntactic representation rather than a superficial orthographic artifact.

\section{Conclusion}

Punctuation plays a structurally meaningful role in shaping syntactic representations, yet conventional binarization procedures often remove it as if it were extraneous. This practice introduces distortions into otherwise well-formed trees and reduces the correspondence between syntactic annotations and their underlying linguistic structure.

This study introduced a linguistically principled restructuring procedure that preserves punctuation as a sibling node prior to binarization. The resulting binary trees remain fully compatible with existing treebanks, require no additional annotation, and can be deterministically reversed to their original form. Empirically, the approach improves head-child identification accuracy and yields structures that align more closely with derivation-based frameworks such as CCGbank.

Beyond its empirical performance, the proposed method reinforces the broader principle that punctuation contributes essential structural cues. Preserving these cues enhances the fidelity of syntactic representations, supports consistent alignment with formal grammar frameworks, and facilitates reproducible preprocessing pipelines across resources.

Although the present work focuses on English, the same principle applies to languages where punctuation marks clause or phrase boundaries, including Chinese and French. Extending punctuation-aware binarization to multilingual settings, potentially through weakly supervised alignment, constitutes a promising direction for future research.

\section*{Data Availability Statement}
The binarization process is available at \url{https://doi.org/10.5281/zenodo.17336455} and is released under the Creative Commons Attribution 4.0 International License.  
The GitHub repository (\url{https://github.com/jungyeul/treebank-tree-binarization}) additionally provides the full implementation, including binarization and head-finding functions, detailed documentation, and experimental results.


\end{document}